# Robust Graphical Modeling with $t$-Distributions

**Michael A. Finegold**
Department of Statistics
The University of Chicago
Chicago, IL 60637

**Mathias Drton**
Department of Statistics
The University of Chicago
Chicago, IL 60637

## Abstract

Graphical Gaussian models have proven to be useful tools for exploring network structures based on multivariate data. Applications to studies of gene expression have generated substantial interest in these models, and resulting recent progress includes the development of fitting methodology involving penalization of the likelihood function. In this paper we advocate the use of the multivariate $t$ and related distributions for more robust inference of graphs. In particular, we demonstrate that penalized likelihood inference combined with an application of the EM algorithm provides a simple and computationally efficient approach to model selection in the $t$-distribution case.

## 1 INTRODUCTION

Graphical Gaussian models have attracted a lot of recent interest. In these models, an observed random vector $Y = (Y_1, \ldots, Y_p)$ is assumed to follow a multivariate normal distribution $\mathcal{N}_p(\mu, \Sigma)$, where $\mu$ is the mean vector and $\Sigma$ the positive definite covariance matrix. Each model is associated with an undirected graph $G = (V, E)$ with vertex set $V = \{1, \ldots, p\}$, and defined by requiring that for each non-edge $(i, j) \notin E$, the variables $Y_i$ and $Y_j$ are conditionally independent given all the remaining variables $Y_{\backslash\{i,j\}}$. Here, $\backslash\{i, j\}$ denotes the complement $V \setminus \{i, j\}$. Such pairwise conditional independence holds if and only if $\Sigma_{ij}^{-1} = 0$. Therefore, inferring the graph corresponds to inferring the non-zero elements of $\Sigma^{-1}$.

Classical solutions to the graph or model selection problem include constraint-based approaches that test the model-defining conditional independence constraints, and score-based searches that optimize a model score over a set of graphs. Recently, however, penalized likelihood approaches based on the one-norm of the concentration matrix $\Sigma^{-1}$ have become increasingly popular. Meinshausen and Bühlmann (2006) proposed a heuristic method that uses lasso regressions of each variable $Y_i$ on the remaining variables $Y_{\backslash i} := Y_{\backslash\{i\}}$. In subsequent work, Yuan and Lin (2007) and Banerjee *et al* (2008) discuss the computation of the exact solution to the convex optimization problem arising from the likelihood penalization. Finally, Friedman *et al* (2008) developed the *graphical lasso (glasso)*, which is a computationally efficient algorithm that maximizes the penalized log-likelihood function through coordinate-descent. The theory that accompanies these algorithmic developments supplies high-dimensional consistency properties under assumptions of graph sparsity; see e.g. Ravikumar *et al* (2008).

Inference of a graph can be significantly impacted, however, by deviations from normality. In particular, contamination of a handful of variables in a few experiments can lead to a drastically wrong graph. Applied work thus often proceeds by identifying and removing such experiments before data analysis, but such outlier screening can become difficult with large datasets. More importantly, removing entire experiments as outliers may discard useful information from the uncontaminated variables they may contain.

The existing literature on robust inference in graphical models is fairly limited. One line of work concerns constraint-based approaches and adopts robustified statistical tests (Kalisch and Bühlmann, 2008). An approach for fitting the model associated with a given graph using a robustified likelihood function is described in Miyamura and Kano (2006).

In this paper we extend the scope of robust inference by providing a tool for robust model selection that can be applied with rather highly multivariate data. We build upon the *glasso* of Friedman *et al* (2008) but model the data using multivariate $t$-distributions. Using the EM algorithm, the *tlasso* we propose is only slightly less computationally efficient than the *glasso*.

In §2 we review maximization of the penalized Gaussian log-likelihood using the *glasso*. In §3 we introduce the multivariate $t$-distribution and review maximization of the (unpenalized) log-likelihood using the EM algorithm. In §4.1 we combine the two techniques into the *tlasso* to maximize the penalized log-likelihood in the multivariate $t$ case; in §4.2 we show simulation results comparing the *glasso* with the *tlasso*; and in §4.3 we use the two methods on gene expression data. Finally, in §5 we show how the *tlasso* can be modified for an alternative multivariate $t$-distribution.

## 2 THE GRAPHICAL LASSO

Suppose we observe a sample of $n$ independent random vectors $Y_1, \ldots, Y_n \in \mathbb{R}^p$ that are distributed according to the multivariate normal distribution $\mathcal{N}_p(\mu, \Sigma)$. Likelihood inference about the covariance matrix $\Sigma$ is based on the log-likelihood function

$$\ell(\Sigma) = -\frac{np}{2}\log(2\pi) - \frac{n}{2}\log\det(\Sigma) - \frac{n}{2}\text{tr}(S\Sigma^{-1}),$$

where the empirical covariance matrix

$$S = \frac{1}{n}\sum_{i=1}^n (Y_i - \bar{Y})(Y_i - \bar{Y})^T$$

is defined based on deviations from the sample mean $\bar{Y}$. Let $\Theta = (\theta_{ij}) = \Sigma^{-1}$ denote the $p \times p$-concentration matrix. In penalized likelihood methods a one-norm penalty is added to the log-likelihood function, which effectively performs model selection because the resulting estimates of $\Theta$ may have entries that are exactly zero. Omitting irrelevant factors and constants, we are led to the problem of maximizing the function

$$\log\det(\Theta) - \text{tr}(S\Theta) - \rho\|\Theta\|_1 \quad (1)$$

over the cone of positive definite matrices. Often the one-norm is defined as

$$\|\Theta\|_1 = \sum_{1 \le i < j \le p} |\theta_{ij}|$$

such that only the off-diagonal entries of $\Theta$ are involved in the regularization term. The multiplier $\rho$ is a positive tuning parameter. Larger values of $\rho$ lead to more entries of $\Theta$ being estimated as zero. The tuning of $\rho$ is typically done through cross-validation or information criteria.

The *glasso* is a computationally efficient method for solving the convex optimization problem in (1). Its iterative updates operate on the covariance matrix $\Sigma$. In each step one row (and column) of the symmetric matrix $\Sigma$ is updated based on a partial maximization of (1) in which all but the considered row (and column) of $\Theta$ are held fixed. This partial maximization is solved via coordinate-descent as briefly reviewed next.

Partition off the last row and column of $\Sigma = (\sigma_{ij})$ and $S$ as

$$\Sigma = \begin{pmatrix} \Sigma_{\backslash p, \backslash p} & \Sigma_{\backslash p, p} \\ \Sigma_{\backslash p, p}^T & \sigma_{pp} \end{pmatrix}, \quad S = \begin{pmatrix} S_{\backslash p, \backslash p} & S_{\backslash p, p} \\ S_{\backslash p, p}^T & S_{pp} \end{pmatrix}.$$

Then, as shown in Banerjee *et al* (2008), partially maximizing $\Sigma_{\backslash p, p}$ with $\Sigma_{\backslash p, \backslash p}$ held fixed yields $\Sigma_{\backslash p, p} = \Sigma_{\backslash p, \backslash p}\tilde{\beta}$, where $\tilde{\beta}$ minimizes

$$\|(\Sigma_{\backslash p, \backslash p})^{1/2}\beta - (\Sigma_{\backslash p, \backslash p})^{-1/2}S_{\backslash p, p}\|^2 + \rho\|\beta\|_1$$

with respect to $\beta \in \mathbb{R}^{p-1}$. The *glasso* finds $\tilde{\beta}$ by coordinate descent in each of the coordinates $j = 1, \ldots, p-1$, using the updates

$$\tilde{\beta}_j = \frac{T\left(S_{jp} - \sum_{k<p, k \ne j}\sigma_{kj}\tilde{\beta}_k, \rho\right)}{\sigma_{jj}}$$

where $T(x, t) = \text{sgn}(x)(|x| - t)_+$. The algorithm then cycles through the rows and columns of $\Sigma$ and $S$ until convergence; see Friedman *et al* (2008).

## 3 GRAPHICAL $T$-MODELS

### 3.1 MULTIVARIATE $T$-DISTRIBUTION

The multivariate $t$-distribution $t_{p,\nu}(\mu, \Psi)$ on $\mathbb{R}^p$ has Lebesgue density

$$f_\nu(y; \mu, \Psi) = \frac{\Gamma(\frac{\nu+p}{2})|\Psi|^{-1/2}}{(\pi\nu)^{p/2}\Gamma(\frac{\nu}{2})[1 + \delta_y(\mu, \Psi)/\nu]^{(\nu+p)/2}} \quad (2)$$

with $\delta_y(\mu, \Psi) = (y - \mu)^T\Psi^{-1}(y - \mu)$ and $y \in \mathbb{R}^p$. The vector $\mu \in \mathbb{R}^p$ and the positive definite matrix $\Psi$ determine the expectation and covariance matrix of the distribution, namely, if $Y \sim t_{p,\nu}(\mu, \Psi)$ with $\nu \ge 3$ degrees of freedom then $\mathbb{E}[Y] = \mu$ and $\mathbb{V}[Y] = \nu/(\nu-2) \cdot \Psi$. From here on we will assume $\nu \ge 3$ such that the covariance matrix exists.

If $X \sim \mathcal{N}_p(0, \Psi)$ is a multivariate normal random vector independent of the Gamma-random variable $\tau \sim \Gamma(\nu/2, \nu/2)$, then $Y = \mu + X/\sqrt{\tau} \sim t_{p,\nu}(\mu, \Psi)$; see Kotz and Nadarajah (2004). This scale-mixture representation, illustrated in Figure 1, allows for easy sampling. It also clarifies how the use of $t$-distributions leads to more robust inference because extreme observations can be generated via small values of $\tau$.

Let $G = (V, E)$ be a graph with vertex set $V = \{1, \ldots, p\}$. We define the associated graphical model for the $t$-distribution by requiring that $\Psi_{ij}^{-1} = 0$ for indices $i \ne j$ corresponding to a non-edge $(i, j) \notin E$.

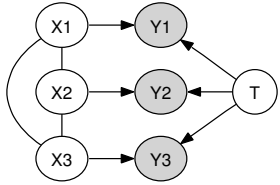

Figure 1: Graph representing the process generating a multivariate $t$-random vector $Y$ from a latent Gaussian random vector $X$ and a single latent Gamma-divisor.

This mimics the Gaussian model in that zero constraints are imposed on the inverse of the covariance matrix. However, in a $t$-distribution this no longer corresponds to conditional independence, and the density $f_\nu(y; \mu, \Psi)$ does not factor according to the graph. The conditional dependence manifests itself in particular in conditional variances in that even if $\Psi_{ij}^{-1} = 0$,

$$\mathbb{V}[Y_i | Y_{\setminus i}] \neq \mathbb{V}[Y_i | Y_{\setminus \{i,j\}}].$$

Nevertheless the following property still holds.

**Theorem 1.** *Let $Y \sim t_{p,\nu}(\mu, \Psi)$, where $\Psi$ is a positive definite matrix with $\Psi_{ij}^{-1} = 0$ for indices $i \neq j$ corresponding to the non-edges in the graph $G$. If two nodes $i$ and $j$ are separated by a set of nodes $C$ in $G$, then $Y_i$ and $Y_j$ are conditionally uncorrelated given $Y_C$.*

This connection to zero conditional correlations is appealing because it entails that mean-square error optimal prediction of variable $Y_i$ can be based on the variables $Y_j$ that correspond to neighbors of the node $i$ in the graph. Alternatively, we can interpret the edges of the graph as indicating the allowed conditional dependencies in the latent Gaussian vector $X$.

### 3.2 EM ALGORITHM FOR ESTIMATION

The lack of density factorization properties complicates likelihood inference with $t$-distributions. However, the EM algorithm provides a way to exploit Gaussian techniques. Equipped with the normal-Gamma construction, we treat $\tau$ as a hidden variable and use that the conditional distribution of $Y$ given $\tau$ is $\mathcal{N}_p(\mu, \Psi/\tau)$. The E-step is simple because

$$\mathbb{E}[\tau | Y] = \frac{\nu + p}{\nu + \delta_Y(\mu, \Psi)}; \quad (3)$$

see Liu and Rubin (1995). We now outline the EM algorithm for the $t$-distribution assuming the degrees of freedom $\nu$ to be known. In practice, $\nu$ could also be estimated in a line search that is best based on the actual $t$-likelihood (Liu and Rubin, 1995).

Consider an $n$-sample $Y_1, \ldots, Y_n$ drawn from $t_{p,\nu}(\mu, \Psi)$. Let $\tau_1, \ldots, \tau_n$ be an associated sequence of hidden Gamma-random variables. Observation of the $\tau_i$ would lead to the following complete-data log-likelihood function for $\mu$ and $\Theta = \Psi^{-1}$:

$$\ell_{\text{hid}}(\mu, \Theta | Y, \tau) \propto \frac{n}{2} \log \det(\Theta) - \frac{1}{2} \text{tr}\left(\Theta \sum_{i=1}^n \tau_i Y_i Y_i^T\right)$$
$$+ \mu^T \Theta \sum_{i=1}^n \tau_i Y_i - \frac{1}{2} \mu^T \Theta \mu \sum_{i=1}^n \tau_i, \quad (4)$$

where, with some abuse, the symbol $\propto$ indicates that irrelevant additive constants are omitted. The complete-data sufficient statistics

$$S_\tau = \sum_{i=1}^n \tau_i, \quad S_{\tau Y} = \sum_{i=1}^n \tau_i Y_i, \quad S_{\tau YY} = \sum_{i=1}^n \tau_i Y_i Y_i^T$$

are thus linear in $\tau$. We obtain the following EM algorithm for computing the maximum likelihood estimates of $\mu$ and $\Psi$:

**E-step:** Given current estimates $\mu^{(t)}$ and $\Psi^{(t)}$ compute $\tau_i^{(t+1)} = (\nu + p)/(\nu + \delta_Y(\mu^{(t)}, \Psi^{(t)}))$.

**M-step:** Calculate the updated estimates

$$\mu^{(t+1)} = \frac{\sum_{i=1}^n \tau_i^{(t+1)} Y_i}{\sum_{i=1}^n \tau_i^{(t+1)}}, \quad (5)$$

$$\Psi^{(t+1)} = \frac{1}{n} \sum_{i=1}^n \tau_i^{(t+1)} [Y_i - \mu^{(t+1)}][Y_i - \mu^{(t+1)}]^T.$$

Note that the E-step uses the result in (3).

## 4 PENALIZED $T$-LIKELIHOOD

### 4.1 THE TLASSO

Model selection in graphical $t$-models can be performed in principle by any of the classical constraint- and score-based methods. In score-based searches through the set of all undirected graphs on $p$ nodes, each model would have to be refit using an iterative method. However, the penalized likelihood approach circumvents this problem in a way that is similar to structural EM algorithms (Friedman, 1997).

Like in the Gaussian case, we put a one-norm penalty on the elements of $\Theta = \Psi^{-1}$ and wish to maximize the penalized log-likelihood function

$$\ell_{\rho, \text{obs}}(\mu, \Theta | Y) = \sum_{i=1}^n \log f_\nu(Y_i; \mu, \Theta^{-1}) - \rho \|\Theta\|_1, \quad (6)$$

where $f_\nu$ is the $t$-density from (2). To achieve this we will use a modified version of the EM algorithm taking into account the one-norm penalty.

We treat $\tau$ as missing data. In the E-step of our algorithm, we calculate the conditional expectation of the penalized complete-data log-likelihood

$$\ell_{\rho,\text{hid}}(\mu, \Theta|Y, \tau) \\ \propto \frac{n}{2}\log|\Theta| - \frac{n}{2}\text{tr}(\Theta S_{\tau YY}(\mu)) - \rho\|\Theta\|_1, \quad (7)$$

where

$$S_{\tau YY}(\mu) = \frac{1}{n}\sum_{i=1}^{n}\tau_i(Y_i - \mu)(Y_i - \mu)^T.$$

Since $\ell_{\rho,\text{hid}}(\mu, \Theta|Y, \tau)$ is again linear in $\tau$, the E-step takes the same form as in §3.2.

Let $\mu^{(t)}$ and $\Theta^{(t)}$ be the estimates after the $t^{th}$ iteration, and $\tau_i^{(t+1)}$ the conditional expectation of $\tau_i$ calculated in the $(t+1)^{th}$ E-step. Then in the M-step of our algorithm we wish to maximize

$$\frac{n}{2}\log|\Theta| - \frac{n}{2}\text{tr}(\Theta S_{\tau^{(t+1)}YY}(\mu)) - \rho\|\Theta\|_1$$

with respect to $\mu$ and $\Theta$. Differentiation with respect to $\mu$ yields $\mu^{(t+1)}$ from (5) for any value of $\Theta$. Therefore, $\Theta^{(t+1)}$ is found by maximizing

$$\frac{n}{2}\log|\Theta| - \frac{n}{2}\text{tr}(\Theta S_{\tau^{(t+1)}YY}(\mu^{(t+1)})) - \rho\|\Theta\|_1. \quad (8)$$

The quantity in (8), however, is exactly the objective function maximized by the *glasso*.

Iterating the E- and M-steps just described we obtain what we call the *tlasso* algorithm. A nice feature of the approach is that convergence guarantees can be given based on the following observation, which can be proven in similar fashion as the corresponding standard result about the ordinary EM algorithm.

**Theorem 2.** *The tlasso never decreases the penalized log-likelihood function of the considered graphical t-distribution model, that is, in all iterations $t = 1, 2, \ldots$ it holds that*

$$\ell_{\rho,\text{obs}}(\mu^{(t+1)}, \Theta^{(t+1)}|Y) \geq \ell_{\rho,\text{obs}}(\mu^{(t)}, \Theta^{(t)}|Y).$$

### 4.2 SIMULATION RESULTS

We ran multiple simulations with a range of $\rho$ values to compare how well the competing methods recovered the true edges. Our *tlasso* is computationally more expensive, since it calls the *glasso* at each M-step. But in our simulations, the algorithm converges quickly. If we run through multiple increasing values of $\rho$, it may take about 15 to 30 EM iterations for the initial small value of $\rho$, but only 2 or 3 iterations for later values as we can start at the previous output. But even in the initial run, two iterations typically lead to a drastic improvement (in the $t$ likelihood) over the *glasso*.

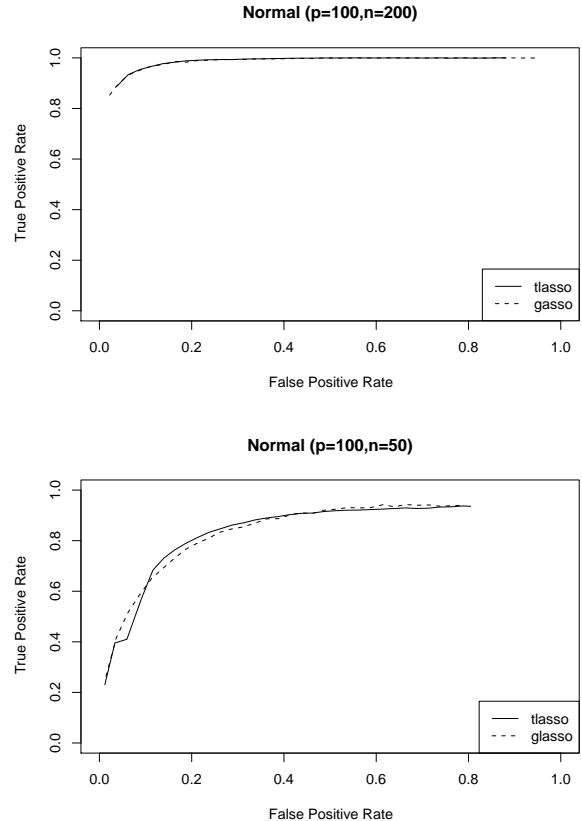

Figure 2: ROC curves based on averaging 50 simulations with $\mathcal{N}_{100}$-data (true concentration matrices have off-diagonal elements non-zero with prob. 0.02). The *tlasso* uses $\nu = 3$ degrees of freedom.

In the worst case scenario for our *tlasso* relative to the *glasso*—when the data is normal and we assume a $t$-distribution with 3 degrees of freedom—almost no statistical efficiency is lost. In the numerous simulations we have run using normally generated data, the *tlasso* and *glasso* do an essentially equally good job of recovering the true graph, as illustrated in Figure 2.

The other extreme occurs for data generated from a $t$-distribution with 3 degrees of freedom. With $p = 100$ nodes, a sparse graph, and $n = 50$ observations, the *tlasso* provides drastic improvement over the *glasso* at the low false positive rates that are of interest in practice (Figure 3). The assumption of normality and the occasional extreme observation lead to numerous false positives when using the *glasso*. With $n = 200$ observations the *tlasso* does a very good job of recovering the true graph and significantly outperforms the normal approach. Therefore, there is very little computational—and little or no statistical—downside to assuming $t$-distributions, but significant statistical upside.

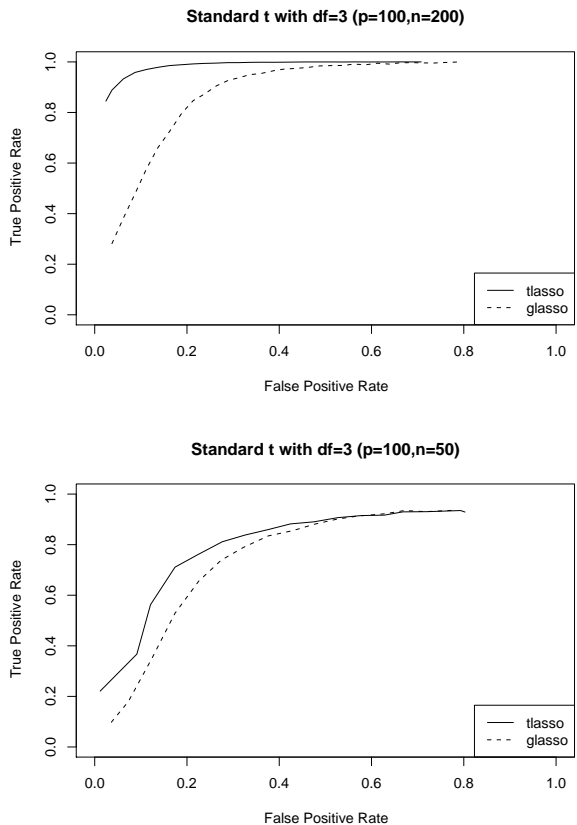

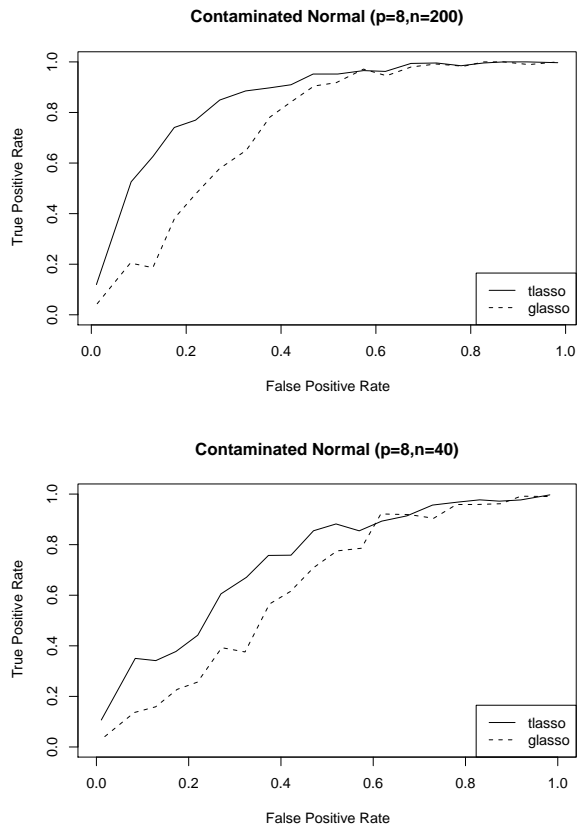

Figure 3: ROC curves based on averaging 50 simulations with $t_{100,3}$-data (true concentrations non-zero with prob. 0.02). The *tlasso* uses $\nu = 3$ df.

Figure 4: ROC curves based on 50 simulations with contaminated $\mathcal{N}_8$-data (true concentrations non-zero with prob. 0.2). In 5% of the observations, the same 3 nodes have data generated instead from univariate normal distributions with mean 25 times the maximum of any variance term in the original multivariate normal. The *tlasso* uses $\nu = 3$ df.

A more realistic setting might be one in which normal data is contaminated in some fashion. Here we assume that 3 nodes out of 8 are contaminated with rather large values in a small portion of the data set. In simulations from such a normal-contamination model, the *tlasso* again outperforms the *glasso*. Even with a large sample size, the latter method tends to obtain false positive edges among the 3 contaminated nodes. The *tlasso* on the other hand downweights the contaminated data points and performs much better.

### 4.3 GENE EXPRESSION DATA

We consider data from microarray experiments with yeast strands (Gasch *et al*, 2000). As in Drton and Richardson (2008), we limit this illustration to 8 genes involved in galactose utilization. An assumption of normality is brought into question in particular by the fact that in 11 out of 136 experiments with data for all 8 genes, the measurements for 4 of the genes were abnormally large negative values. In order to assess the impact of such a handful of outliers, we run each algorithm, adjusting the penalty term $\rho$ such that a graph with a given number of edges is inferred. Somewhat arbitrarily we will focus on the top 9 edges. We do this once with all 136 experiments and then again excluding the 11 potential outliers.

As seen in Figure 5, the *glasso* infers very different graphs, with only 3 edges in common. When the "outliers" are included, the *glasso* estimate in Fig. 5(a) has the 4 nodes in question fully connected; when they are excluded, no edges among the 4 nodes are inferred. The *tlasso* does not exhibit this extreme behavior. As shown in Fig. 5(b) it recovers almost the same graph in each case (7 out of 9 edges shared). When run with all the data, the $\tau$ estimate is very small ($\sim 0.04$) for each of the 11 questionable observations compared with the average $\tau$ estimate of 1.2. The graph in Fig. 5(c) shows the results from the *alternative tlasso* discussed in §5, which performs just as well as the *tlasso*.

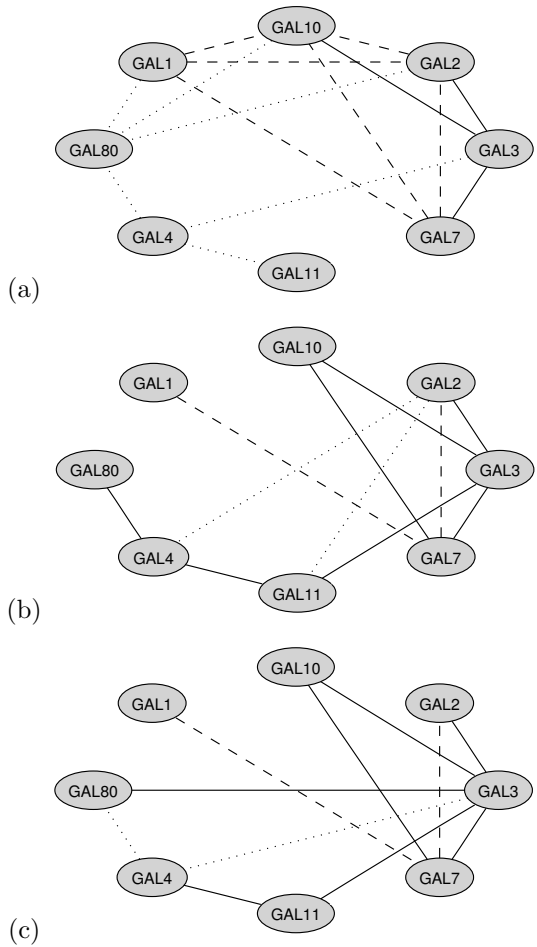

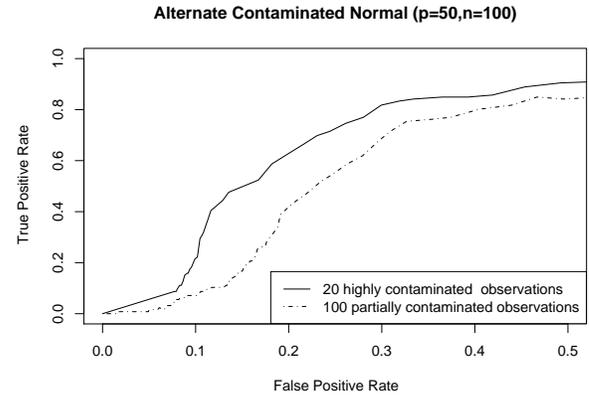

Figure 6: ROC curves for contaminated $\mathcal{N}_{50}$-data (true concentrations non-zero with prob. 0.1). First, the same 15 nodes are contaminated in 20 observations. Second, for 5 blocks of 20 observations a different set of 3 nodes is contaminated. All "contaminated" nodes are drawn from univariate normals with mean 10 times the maximum of any covariance. The *tlasso* uses $\nu = 3$ df.

Figure 5: Top 9 recovered edges: (a) *glasso*, (b) *tlasso*, (c) alternative *tlasso* discussed in §5. Dashed edges were recovered only when including the outliers; dotted only when excluding them; solid in both cases.

## 5 ALTERNATIVE MODEL

### 5.1 MOTIVATION

The *tlasso* from §4 performs particularly well when a small fraction of the observations are contaminated. In this case, these observations are downweighted in entirety and the gain from reducing the effect of contaminated nodes outweighs the loss from throwing away good data from the other nodes. In very high-dimensional datasets, however, the contamination may be in small parts of many observations. Downweighting entire observations may then no longer achieve the desired results.

In the illustrative example below, we evaluate our methods on two types of data. The same underlying $\mathcal{N}_{50}$ distribution is used to generate the data in both cases. In the first case, represented by the top curve, the same 15 nodes are contaminated in 20 observations; in the second case, a different group of 3 nodes is contaminated in each of 5 blocks of 20 observations. The overall contamination level, concentration matrix, and uncontaminated data are exactly the same for both cases, but the difference in the performance of the *tlasso* in the two cases is significant. In the first case, the *tlasso* downweights the contaminated observations and uses the rest of the data to recover the graph. In the second case, all observations are contaminated in some coordinates and there is little gained from weighting them differently. The *tlasso* cannot cope with this latter situation and will generate false positives for the edges connecting the co-contaminated nodes.

The results of the *glasso* are not shown in the figure for the sake of clarity, but the *tlasso* still outperforms the *glasso* in both cases. In the second case, the *tlasso* at least downweights the most extreme cases. The normal model obtains false positives for edges for almost all of the co-contaminated nodes, except for values of $\rho$ so small that most of the true positives are missed.

### 5.2 SPECIFICATION OF THE ALTERNATIVE $T$-MODEL

To handle the above situation better, we consider an alternative extension of the univariate *t*-distribution, illustrated in Figure 7. Instead of one divisor $\tau$ per *p*-variate observation, we draw $p$ divisors $\tau_j$. For $j = 1, \ldots, p$, let $\tau_j \sim \Gamma(\nu/2, \nu/2)$ be independent of each other and of $X \sim \mathcal{N}_p(0, \Psi)$. We then say that the random vector $Y$ with coordinates $Y_j = \mu_j + X_j/\sqrt{\tau_j}$ follows an alternative multivariate $t$-distribution. We will denote this as $Y \sim t^*_{p,\nu}(\mu, \Psi)$.

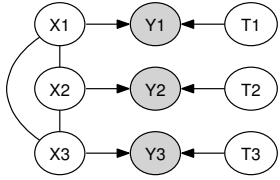

Figure 7: Graph representing the process generating a $t^*$-random vector $Y$ from a latent Gaussian random vector $X$ and independent latent Gamma-divisors.

Unlike for the standard multivariate $t$-distribution, the covariance matrix $\mathbb{V}[Y]$ is no longer a constant multiple of $\Psi$ when $Y \sim t^*_{p,\nu}(\mu, \Psi)$. Clearly, the coordinate variances are still the same, namely $\mathbb{V}[Y_i] = \nu/(\nu-2) \cdot \psi_{ii}$, but the covariance between $Y_i$ and $Y_j$ with $i \neq j$ is now

$$\frac{\nu \Gamma(\frac{\nu-1}{2})^2}{2\Gamma(\frac{\nu}{2})^2} \cdot \psi_{ij} \leq \frac{\nu}{\nu-2} \cdot \psi_{ij}.$$

The same matrix $\Psi$ thus implies smaller correlations (by the same constant multiple) in the $t^*$-distribution. This reduced dependence is not surprising in light of the fact that now different and independent divisors appear in the different coordinates. Despite the decrease in marginal correlations, the result of Theorem 1 does not appear to hold for conditional correlations in the alternative model.

### 5.3 ALTERNATIVE TLASSO

Inference in the alternative model presents some difficulties because the likelihood is not available explicitly. The complete-data log-likelihood function $\ell^*_{\rho,\text{hid}}(\mu, \Theta | Y, \tau)$, however, is simply the product of the evaluations of $p$ Gamma-densities ($\tau$ being a vector now) and a multivariate normal density. We can thus implement an EM-type procedure if we are able to compute the conditional expectation of $\ell^*_{\rho,\text{hid}}(\mu, \Theta | Y, \tau)$ given $Y = (Y_1, \ldots, Y_n)$. This time we treat the $p$ random variables $(\tau_{i1}, \ldots, \tau_{ip})$ as hidden for each observation $i = 1, \ldots, n$. Unfortunately, the conditional expectation is intractable. However, it can be estimated using Markov Chain Monte Carlo.

The complete-data log-likelihood function is equal to

$$\ell^*_{\rho,\text{hid}}(\mu, \Theta, |Y, \tau)$$
$$\propto \frac{n}{2} \log |\Theta| - \frac{n}{2} \text{tr}\big(\Theta S^*_{\tau YY}(\mu)\big) - \rho \|\Theta\|_1 \quad (9)$$

where

$$S^*_{\tau YY}(\mu) = \frac{1}{n} \sum_{i=1}^n D(\sqrt{\tau_i})(Y_i - \mu)(Y_i - \mu)^T D(\sqrt{\tau_i})$$

and $D(\sqrt{\tau_i})$ is the diagonal matrix with $\sqrt{\tau_{i1}}, \ldots, \sqrt{\tau_{ip}}$ along the diagonal. The trace in (9) is linear in the entries of the matrix $\sqrt{\tau_i}\sqrt{\tau_i}^T$. A Markov Chain Monte Carlo procedure for estimating the conditional expectation of this matrix given $Y$ cycles through the coordinates indexed by $j = 1, \ldots, p$ and accepts or rejects a draw from a proposal distribution in order to sample from the conditional distribution of $\tau_{ij}$ given $(\tau_{i\setminus j}, Y)$. This full conditional is proportional to

$$q(\tau_{ij}) \exp \left\{ -\tau_{ij}^{1/2}(Y_{ij} - \mu_j)\Theta_{j\setminus j} X_{i\setminus j} \right\}$$

where $q(\tau_{ij})$ is the density of the Gamma-distribution:

$$\Gamma\left(\frac{\nu+1}{2}, \frac{\nu + (Y_{ij} - \mu_j)^2 \theta_{jj}}{2}\right)$$

We then use $q(\tau_{ij})$ as the proposal density. We calculate $\sqrt{\tau_i}\sqrt{\tau_i}^T$ at the end of each cycle through the $p$ nodes, and then take the average over $K$ iterations. This solves the problem of carrying out one E-step, and we obtain the following EM-like algorithm, which we call the *alternative tlasso*:

**E-step:** Given current estimates $\mu^{(t)}$ and $\Psi^{(t)}$ compute $\left(\sqrt{\tau_i}\sqrt{\tau_i}^T\right)^{(t+1)} = \frac{1}{K} \sum_{k=1}^K \left(\sqrt{\tau_i}\sqrt{\tau_i}^T\right)^{(k)}$.

**M-step:** Calculate the updated estimates

$$\mu_j^{(t+1)} = \frac{\sum_{i=1}^n \tau_{ij}^{(t+1)} Y_{ij}}{\sum_{i=1}^n \tau_{ij}^{(t+1)}}.$$

Use these and $\left(\sqrt{\tau_i}\sqrt{\tau_i}^T\right)^{(t+1)}$ to compute the matrix $S^*_{\tau^{(t+1)}YY}(\mu^{(t+1)})$ to be plugged into the trace term in (9). Maximize the resulting penalized log-likelihood function using the *glasso*.

### 5.4 GENE EXPRESSION DATA

With the gene expression data from §4.3, the *alternative tlasso* performs similarly to the *tlasso* when it comes to graph recovery. However, it appears to make better use of the available data. As shown in Figure 8, it significantly downweights the 11 potential outlier observations for the 4 nodes in question, but not for the other nodes. Thus the alternative version is able to extract information from the "uncontaminated" part of the 11 observations while downweighting the rest.

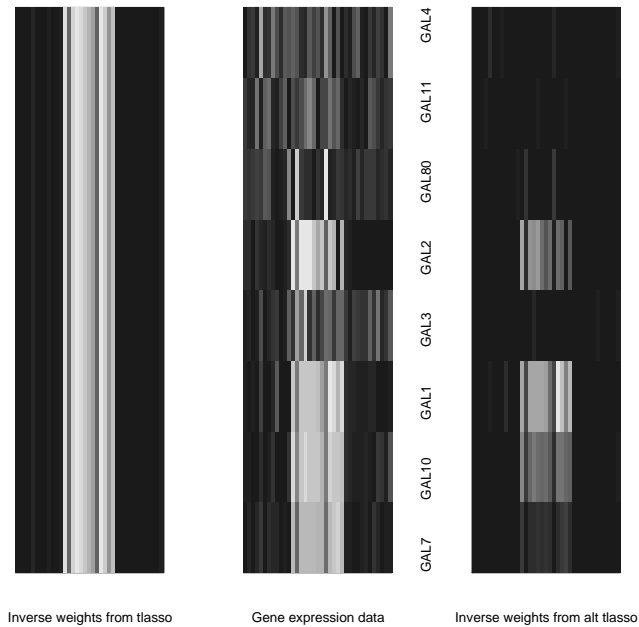

Figure 8: Normalized gene expression data (center), and inverse weights from *tlasso* (left) and *alternative tlasso* (right). Rows correspond to genes and colums to observations. Lighter shades indicate larger values. The *tlasso* uses only one weight per observation and so must weight each gene the same. All plots show the same subset of data including 11 potential outliers.

## 6 DISCUSSION

Our proposed *tlasso* algorithm is a simple and effective method for robust inference in graphical models. Only slightly more computationally expensive than the *glasso*, it can offer great gains in statistical efficiency. The *alternative tlasso* is a more flexible extension. We currently use a Markov Chain Monte Carlo sampler to carry out a stochastic E-step in this alternative procedure. This approach is computationally expensive, and we are in the process of exploring variational approximations.

We assumed $\nu = 3$ degrees of freedom in our various *tlasso* runs. As suggested in prior work on $t$-distribution models, estimation of the degrees of freedom can be done efficiently by a line search based on the observed log-likelihood function. For the *alternative tlasso*, we could employ the hidden log-likelihood function. Nevertheless, our own experience and related literature suggest that not too much is lost by fixing the degrees of freedom at some small value.

Finally, we remark that other normal scale-mixture models could be treated in a similar fashion as the $t$-distribution models we considered in this paper. However, the use of $t$-distributions is particularly convenient in that it is rather robust to various types of misspecification and involves only the single degrees of freedom parameter for the distribution of Gamma-divisors.

### Acknowledgments

This material is based upon work supported by the National Science Foundation (DMS-0746265). M.D. also acknowledges support from an Alfred P. Sloan Research Fellowship.